\documentclass[default,iicol]{sn-jnl}

\jyear{2023}%

\theoremstyle{thmstyleone}%
\theoremstyle{thmstyletwo}%

\theoremstyle{thmstylethree}%

\begin{document}

\title[Article Title]{SAKA: An Intelligent Platform for Semi-automated Knowledge Graph Construction and Application}


\author[1]{\fnm{Hanrong} \sur{Zhang}}\email{hanrong.22@intl.zju.edu.cn}
\equalcont{These authors contributed equally to this work.}

\author[1]{\fnm{Xinyue} \sur{Wang}}\email{xinyue1.22@intl.zju.edu.cn}
\equalcont{These authors contributed equally to this work.}


\author[2]{\fnm{Jiabao} \sur{Pan}}\email{3200102835@zju.edu.cn}

\author*[1]{\fnm{Hongwei} \sur{Wang}}\email{hongweiwang@intl.zju.edu.cn}

\affil[1]{\orgdiv{Zhejiang University—University of Illinois at Urbana-Champaign Joint Institute}, \orgname{Zhejiang University}, \orgaddress{\city{Haining}, \postcode{314400}, \state{Zhejiang}, \country{China}}}
\affil[2]{\orgdiv{Chu Kochen Honors College}, \orgname{Zhejiang University}, \orgaddress{\city{Hangzhou}, \postcode{310058}, \state{Zhejiang}, \country{China}}}


\abstract{Knowledge graph (KG) technology is extensively utilized in many areas, and many companies offer applications based on KG. Nonetheless, the majority of KG platforms necessitate expertise and tremendous time and effort of users to construct KG records manually, which poses great difficulties for ordinary people to use. Additionally, audio data is abundant and holds valuable information, but it is challenging to transform it into a KG. What's more, the platforms usually do not leverage the full potential of the KGs constructed by users. 
In this paper, we propose an intelligent and user-friendly platform for Semi-automated KG Construction and Application (SAKA) to address the problems aforementioned. Primarily, users can semi-automatically construct KGs from structured data of numerous areas by interacting with the platform, based on which multi-versions of KG can be stored, viewed, managed, and updated. Moreover, we propose an Audio-based KG Information Extraction (AGIE) method to establish KGs from audio data. 
Lastly, the platform creates a semantic parsing-based knowledge base question answering (KBQA) system based on the user-created KGs. We prove the feasibility of the semi-automatic KG construction method on the SAKA platform.}




\maketitle

\section{Introduction}\label{sec1}

The rise of big data in recent years have posed significant difficulties in managing, processing and understanding vast amounts of data. Knowledge graph (KG), as a graph-based storing utility, encodes facts amongst various entities (nodes or ontologies), which offers a novel method to better arrange vast data. Based on the rapid development of machine learning technologies~\cite{li-etal-2023-class,10114639,peng2023sclifdsupervisedcontrastiveknowledgedistillation,10152774,xie2023empirical}, KG is increasingly prevalent and has been applied to numerous fields, such as media and geography.

Despite the potential benefits of KG, most KG platforms are complex and demand specialized expertise to use correctly. Constructing KGs manually requires significant time and effort, and this process is usually beyond the capabilities of the average user. This poses a significant challenge for many individuals who are interested in using KGs for various purposes but lack the necessary skills and resources. Furthermore, while audio data is a valuable source of information, it's often challenging to transform this data into a format that is usable in a KG. This is because audio data can be difficult to structure and analyze, and the conversion process can be time-consuming and cumbersome.
Finally, while many KG platforms allow users to create their own KGs, many of these platforms do not fully utilize the potential of the KGs created by users, limiting the usefulness and impact of these resources~\cite{zhang2022intelligent}. 

\begin{figure}[ht]
    \centering
    \includegraphics[width=0.48\textwidth]{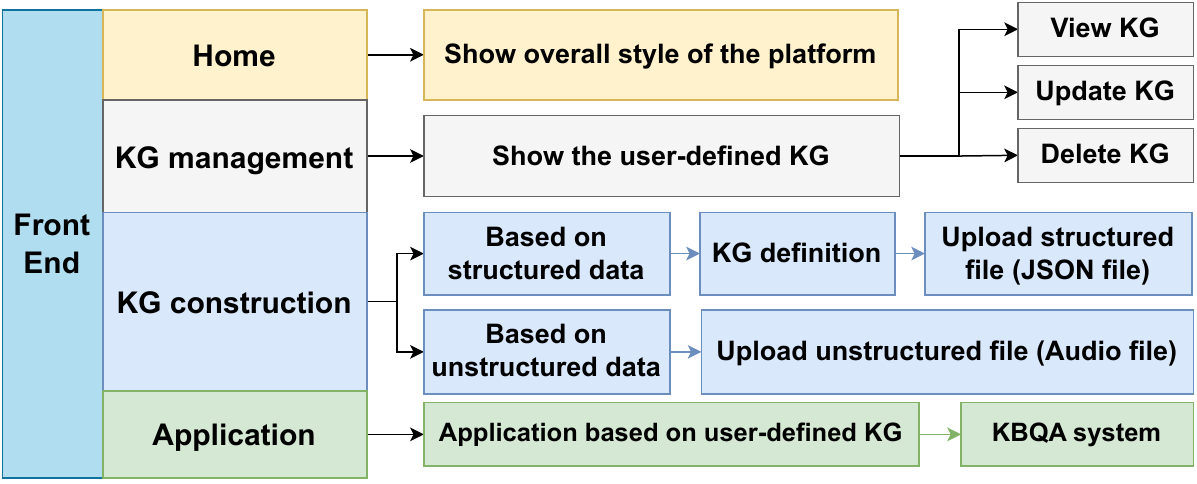}
    \caption{Basic Architecture of SAKA}
    \label{BArchitecture}
\end{figure}

In this article, to tackle the problems mentioned above, we propose an intelligent platform for Semi-automated KG Construction and Application (SAKA), which mainly consists of the following components: the KG construction module, the KG management module, and the application module. Primarily, knowledge about entities is acquired by incorporating data from several structured databases or data records obtained from unstructured data \cite{weikum2021machine}. KG construction refers to the process of cleaning, merging, and blending the data into an exact and consistent representation for each entity.
Therefore, the KG construction component is comprised of KG definition, structured file-based, and audio file-based KG construction. 
Moreover, constant updating of information is crucial because access to up-to-date and reliable  knowledge plays a prominent role in KG construction and application.
Accordingly, the KG management component can also view, modify, and remove these KGs. 
Finally, traditional search engines usually return related web pages instead of the most straightforward answer, where users may need to search again to obtain the final answer. Contrary to traditional search engines, a knowledge base question answering (KBQA) system can answer natural language queries directly using a knowledge base (KB) as the source of knowledge. As a result, the application component implements a semantic parsing-based (SP-based) KBQA module based on the KG constructed by the user as a KB. 
We crawl a medical website for the KB data collection to illustrate the function.
The general structure of the platform is illustrated in Fig. \ref{BArchitecture}.

In conclusion, our main contributions are summarized as follows:
\begin{enumerate}
    \item We create a user-friendly and interactive platform SAKA for intelligent KG construction and application. Users can first customize their desired KG by defining the entity types and relations. Then the user can construct the KG by uploading a structured data file. Multi-domain KGs can be constructed by a file of a defined format and multi-version KGs can be stored, managed, and modified by users.
    \item We propose an Audio-based KG Information Extraction (AGIE) method to establish KGs by semi-automatically extracting semantic information in audio on the KG platform. The entities and relations can be extracted by Voice Activity Detection (VAD) and Speaker Diarization (SD), and Medical Information Extractor (MIE) model. 
    \item A SP-based KBQA module of the medical field is accomplished based on a KB created by users. It can transform users' questions into queries to the KG and provide accurate answers.
    \item We prove the feasibility of the semi-automatic KG construction method on the SAKA platform. Moreover, we evaluate the effectiveness of the VAD, SD, and MIE module of AGIE on the LibriSpeech, VoxCeleb, and doctor-patient dialogue datasets respectively, and achieve decent performance.
\end{enumerate}

\section{Related work}



\subsection{KG construction platform}
Currently, there are numerous KG construction applications to build KGs. Neo4j is one of the popular KG construction and graph database platforms, which constructs KGs by structured data and Cypher statements. 
It requires users to master code capability and Cypher syntax.
Moreover, A deep learning based traditional Chinese medicine knowledge graph platform (TCMKG) is proposed to build KGs from traditional Chinese medicine in 
an end-to-end mode, which also provides knowledge retrieval, visualization, and management functions \cite{Zheng_Liu_Zhang_Wen_2020}. 
Ilyas et al. introduce a knowledge construction and service platform named Saga, which integrates a vast amount of entities in the real world and constructs a central KG that supports numerous production use cases \cite{Ilyas_Rekatsinas_Konda_Pound_Qi_Soliman_2022}.

\subsection{Construction of KG}
To construct a KG, entity extraction is performed first, followed by extracting relationships among entities to realize the construction of a KG. For entity extraction, Huang et al. used Bi-Long Short-Term Memory (Bi-LSTM) model combined with Conditional Random Field (CRF) model based on the word and phrase chunking annotation method to achieve good results in entity recognition tasks \cite{huang2015bidirectional}. Later, the emergence of the Bidirectional Encoder Representations from Transformers (BERT) model greatly influenced subsequent research on entity extraction tasks, verifying the feasibility of pre-trained models for entity extraction tasks and providing a new idea for entity extraction techniques \cite{devlin2019bert}. For relationship extraction, the relationship extraction model proposed in \cite{wang-etal-2016-relation} introduces an attention mechanism based on a convolutional neural network (CNN) for better extraction results. To reduce manual labeling costs, Mintz et al. proposed remote supervision for automatic labeling \cite{mintz-etal-2009-distant} After the BERT model was proposed, the model was widely used in the relational extraction task and was found to be better than CNN and Attention-CNN in the relational extraction task after experiments of Wu et al.\cite{wu2019enriching}.

\subsection{KBQA systems}
Early work on KBQA concentrates on addressing a basic question with a single fact \cite{bordes2015large}. In recent years, researchers have begun to focus more on addressing complicated queries about KBs, i.e., the complex KBQA task \cite{hu_etal_2018_state}. There are two mainstream approaches proposed to address the simple KBQA: SP-based methods and information retrieval-based methods (IR-based methods) \cite{bordes2015large}. SP-based methods represent a query in a format of symbolic logic, which is then executed against the KB to acquire the final answers. They typically follow a chain of modules that includes question comprehension, logical parsing, KB grounding, and KB execution \cite{diefenbach_lopez_singh_maret_2018}.
IR-based methods build a question-related KG that delivers comprehensive knowledge about the question, to which all entities in the retrieved KG are rated according to their importance. The two kinds of methods first identify the subject of a query and associate it with a topic entity in the KB. Next, they either run a parsed logic form or reasoning in a question-related graph to deduce the answers within the neighborhood of the topic entity \cite{Lan_He_Jiang_Jiang_Zhao_Wen_2021}.

\section{Methodology}
\subsection{KG Construction Based on Structured Data}
\begin{figure}[htbp]
    \centering
    \includegraphics[width=\linewidth]{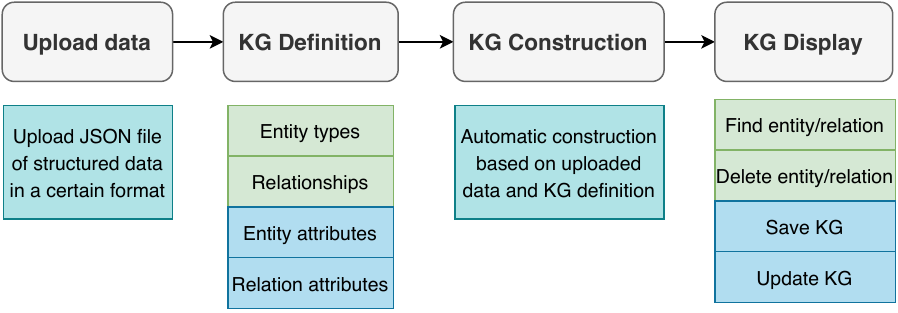}
    \caption{Procedures of KG Construction Based on Structured Data}
    \label{bsd}
\end{figure}

The KG construction process by structured data is split into the following four procedures: 
\begin{enumerate}
    \item Upload a JSON file containing the structured data in a certain format
    \item Define the user-desired KG, which consists of entity types, relationships, entity attributes, and relationship attributes
    \item KG automatic construction based on the uploaded data and KG definition.
    \item Display the constructed KG, which can be searched, modified, and stored in the database.
\end{enumerate}
The KG construction procedures flow is shown in Fig. \ref{bsd}. 
Users are required to operate the module under certain rules to construct the KG in the first two steps, which we will elaborate on in the next subsections. After that, the construction and the display of the KG are done automatically by the backend server. 

\subsubsection{Definition of Uploaded Data}
Initially, the user should upload a JSON file containing all the entities belonging to the topic entity type $O_0$ by numerous data entries. The format of each data entry can be defined as follows:
\begin{equation*}
\{O_0:e,R_i:E_i,A_j:Attr_j\}, i=1 \cdots n, j=1 \cdots m
\end{equation*}
where $O_0$ denotes the topic entity type, and $e$ denotes the entity of the topic entity type of the data entry. $R_i$ denotes the relationships between the topic entity type and other entity types. $A_j$ denotes the attributes of $O_0$, of which $Attr_j$ denotes the attribute content. $E_i$ denotes the entity set of other entity types, which can be defined as:
\begin{equation*}
E_i=\{e_{i_0},e_{i_1},\cdots,e_{i_k}\}
\end{equation*}
where $k \geq 0$ and $k$ are not fixed.

\subsubsection{KG Definition}
\label{definition}
After uploading the data file to the system, the user should manually define the entity types, relationships, entity attributes, and relation attributes contained in the KG, which are also corresponding to the data file format defined above.
Initially, the entity types and relationships should be defined. The entity types consist of the topic entity type $O_0$ and other entity types $O_1-O_n$. This part should also define the relationships of $R_i$ between the topic entity type and other entity types. 
Next, entity attributes and relation attributes need to be defined. The attributes here must have appeared as keys in the JSON data, i.e., one of $A_j$, so that the corresponding attribute values can be acquired when constructing the KG.

\subsubsection{KG Construction}
\begin{figure}[htbp]
    \centering
    \includegraphics[width=\linewidth]{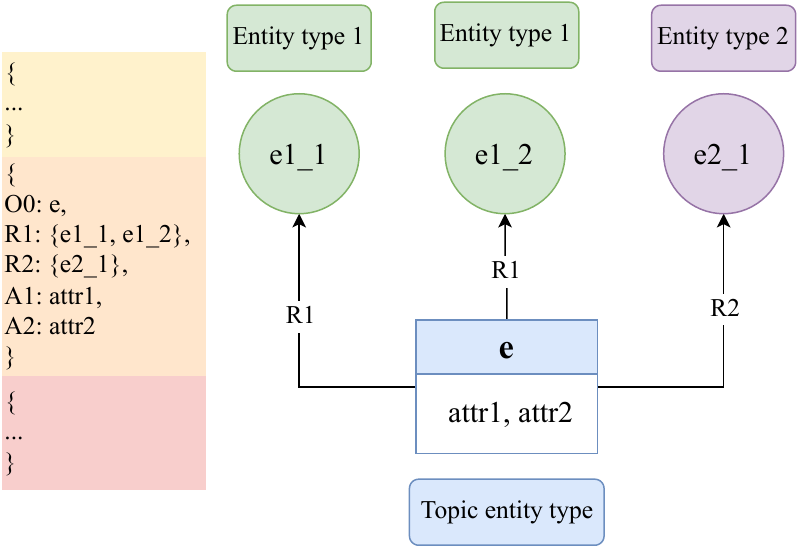}
    \caption{Mapping process between defined KG and JSON data}
    \label{eor}
\end{figure}

After the KG definition, the definition information and KG data file will be uploaded to the back-end server and the subsequent KG construction starts. Initially, the defined entity types, relationships, and attributes are mapped to the uploaded data file in JSON format to facilitate KG construction. A mapping process example is illustrated in Fig. \ref{eor}. The topic entity type is $O_0$, whose attributes $A_1$ and $A_2$ are mapped to the entity $e$ of the topic entity type. $e$ is in relation 1 to $e_{1_1}$ and $e_{1_2}$ of entity type 1 and in relation 2 to $e_{2_1}$ of entity type 2.

Next, the defined KG and KG data will be aggregated into a data dictionary, which consists of ``nodeList'' and ``lineList'' variables. ``nodeList'' is comprised of the definition information of all entity types, including entity type name, and entity attributes; ``lineList'' is comprised of the definition information of all relationships, including relation name, relational entity types pairs, and relation attributes. 

The graph database utilized for KG construction is Neo4j, which utilizes Cypher statements to store information in the form of KGs. The algorithms details are shown in the Algorithm \ref{create_nodes}, \ref{create_rels}.
Initially, the KG definition information in the section \ref{definition} is transformed into two variables to store the KG definition: ``nodeList'' is transformed into ``entityList'', consisting of all the entities in each entity type; ``lineList'' is transformed into ``relationList'', consisting of all the relationships between different entities. Next, the entities and relationships are constructed according to the ``entityList'' and ``relationList''. First, the ``entityList'' and ``relationList'' are de-duplicated. Then the nodes are created by iterating over the key-value pairs in the ``entityList''. After that, the relationships between the nodes are connected according to the ``relationList''.

    

    
    
	
        


       
       



    
    

\begin{algorithm}
\label{create_nodes}
\caption{Create graph nodes}
\begin{algorithmic}
\Procedure{CreateGraphNodes}{}
\State \textbf{Input:} Node\_infos, entity\_list, id\_name
\State \textbf{Output:} Graph nodes

\State \Call{CreateTopicNodes}{Node\_infos} 

// Create topic entity nodes
\State \Call{RemoveDuplicates}{} 

// Remove duplicates from the entity list

\For{each key,value in entity\_list}

// Iterate over entity list and create nodes for other entities
    \State \Call{CreateNode}{id\_name[key], value} // Create other nodes using Cypher statements
\EndFor

\EndProcedure
\end{algorithmic}
\end{algorithm}

\begin{algorithm}
\label{create_rels}

\caption{Create graph relationships}
\begin{algorithmic}
\Procedure{CreateGraphRels}{}
\State \textbf{Input:} relation\_list, relation\_dict, id\_name, id\_relation, id\_transname
\State \textbf{Output:} Graph relationships

\For{each key, value in relation\_list} 

// Iterate over the relation list and create relationships between entities
    \State start\_id $\gets$ relation\_dict[key]['from']
    \State start\_name $\gets$ id\_name[start\_id]
    \State end\_id $\gets$ relation\_dict[key]['to']
    \State end\_name $\gets$ id\_name[end\_id]
    \State rel\_type $\gets$ id\_relation[key]

    \State \Call{CreateRelationship}{start\_name, end\_name, value, rel\_type} 
    
    // Create relationships using Cypher statements
\EndFor

\EndProcedure
\end{algorithmic}
\end{algorithm}

\subsection{KG Construction Based on Audio}
\subsubsection{The AGIE method}

Despite the structured data-based method, 
we propose the AGIE method to establish KGs based on audio on the KG platform. The AGIE method implements audio-preprocessing algorithms to distinguish the speakers in the audio and convert the audio segments into text. Then, the proposed method uses the MIE model \cite{zhang-etal-2020-mie} to extract entities and relations from the dialogue to generate the KG.

\subsubsection{Audio Preprocessing}

\begin{figure}[htbp]
    \centering
    \includegraphics[width=\linewidth]{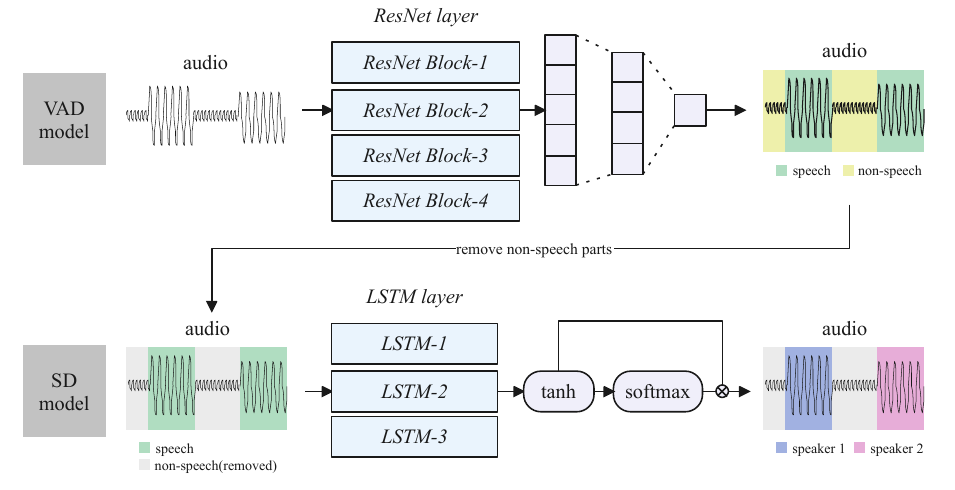}
    \caption{The architecture of VAD model and GE2E model}
    \label{ge2e}
\end{figure}

There are two steps of audio preprocessing. First, imply the VAD model removes the non-speech parts of the audio, and then the SD model is used to find the speaker segmentation points. 
The VAD model uses the ResNet network to train MFCC features of the audio data and classify speech and non-speech segments. After eliminating the non-speech section with the VAD model, the method uses the SD model to identify speakers in the dialogue. Based on the GE2E model \cite{8462665}, the proposed SD method can generate the d-vector of the audio \cite{6854363}, which represents the feature map of the audio data. The GE2E model is structured with multi-layer LSTM. During training, the model obtains 40-mel Filterbank feature from the audio to learn the d-vector feature map. The structures of the proposed models are presented in Fig. \ref{ge2e}.

\subsubsection{Relationship Extraction Model}

With the preprocessed-audio clip, the method can convert audio to text. Then, the relation extraction model is applied to extract information from the converted dialogue. In this paper, we applied the Medical Information Extractor (MIE) model \cite{zhang-etal-2020-mie} as the information extractor. The MIE model is trained to capture critical information in dialogues, with Bi-LSTM layers \cite{6795963} and attention mechanism \cite{NIPS2017_3f5ee243} to learn the time-series-based dialogue information and emphasize the keyword in the conversation. 

The MIE model has four modules: encoder module, matching module, aggregation module, and scoring module.


The encoder module obtains bi-directional features based on Bi-LSTM layers. Then, the module implies the attention mechanism to learn features. Given a statement ${X}_{i}=\lbrace{x}_{1}, {x}_{2},...,{x}_{n}\rbrace$, the network outputs $F[i]$ and the attention weight $a$ to produce the final output $d$, as shown in equation \ref{www}.
\begin{equation}
\begin{aligned}
a[i]&=softmax(W \cdot F[i]+b) \\
d&=\sum_{i=1}^n a[i] \cdot F[i]  
\end{aligned}
\label{www}
\end{equation} 
The $Encoder$ generates two final outputs: $F$ and $d$. $F$ is produced by the Bi-LSTM layer, which denotes the bi-directional features of statement $X$ itself. $d$ represents the feature of $X$ weighted by the attention weight $a$.

The labels of dialogue $D=\{ D[1], D[2],..., D[k]\}$ are categorized into three types:  Category, Item, and Status, which represent the mentioned situation of a patient in the dialogue. Each label is formatted as ``Category: Item-Status'' to describe the information of the three combined categories, where  ``Category: Item'' is regarded as the $c$ part and ``Status'' is called the $s$ part. The $c$ and $s$ part has their corresponding output $F$ and $d$, which is shown in equation \ref{c} and \ref{s}.

\begin{equation}
    {F}^{D}_{c}[i],{d}^{D}_{c}[i]=Encoder^{D}_{c}(D[i])
    \label{c}
\end{equation}
\begin{equation}
    {F}^{D}_{s}[i],{d}^{D}_{s}[i]=Encoder^{D}_{s}(D[i])
    \label{s}
\end{equation}

Like the ${Encoder}^{D}$ for dialogues, the ${Encoder}^{L}$ for labels $L$ is defined in equation \ref{lc} and \ref{ls}.

\begin{equation}
    {F}^{L}_{c}[i],{d}^{L}_{c}[i]=Encoder^{L}_{c}(c)
    \label{lc}
\end{equation}
\begin{equation}
    {F}^{L}_{s}[i],{d}^{L}_{s}[i]=Encoder^{L}_{s}(s)
    \label{ls}
\end{equation}

 The method has four different types of encoders to encode the dialogues and labels in label $c$ and label $s$: ${Encoder}^{D}_{c}$ and ${Encoder}^{D}_{s}$ are encoders for dialogues. ${Encoder}^{L}_{c}$ and ${Encoder}^{L}_{s}$ are encoders for labels.

 
 After obtaining the output $F$ and $c$ for both dialogue and labels, the dialogue and its corresponding label can be matched based on the attention mechanism, as shown in equation \ref{matchC1} and \ref{matchC2}.
 \begin{equation}
     {a}_{c}[i,j]=softmax({d}^{L}_{c},{F}^{D}_{c}[i,j])
     \label{matchC1}
 \end{equation}
 \begin{equation}
     {q}_{c}[i]=\sum_{j} {a}_{c}[i,j] \cdot {F}^{D}_{c}[i,j]
     \label{matchC2}
 \end{equation}
 
  where ${D}[i,j]$ denote the ${j}^{th}$ word of the ${i}^{th}$ sentence in the dialogue. The output ${F}^{D}_{c}$ represents the dialogue within the ``Category: Item-Status'' information. The output ${d}^{L}_{c}$ represents the corresponding ``Category: Item-Status'' label. Thus, ${F}^{D}_{c}$ and ${c}^{L}_{c}$ are matched in this module. Similarly, the output ${F}^{D}_{s}$ generated from dialogues with ``Status'' part information is also matched to ${d}^{L}_{s}$, as shown in equation \ref{matchS1} and \ref{matchS2}.
  
  \begin{equation}
     {a}_{s}[i,j]=softmax({d}^{L}_{s},{F}^{D}_{s}[i,j])
     \label{matchS1}
 \end{equation}
 \begin{equation}
     {q}_{s}[i]=\sum_{j} {a}_{s}[i,j] \cdot {F}^{D}_{s}[i,j]
     \label{matchS2}
 \end{equation}


In the aggregation module, the ${q}_{c}[i]$ and ${q}_{s}[i]$ of the are concatenated to form the full ``Category: Item-Status'' information, as presented in equation \ref{aggre}.
 \begin{equation}
     f[i]=concat({q}_{c}[i],{q}_{s}[i])
     \label{aggre}
 \end{equation}
where $f[i]$ contains the fused information of ``Category: Item'' part $c$ and ``Status'' part $s$, which can predict the score.


In the scoring module, the method uses $f[i]$ to calculate the final output score. The highest score in all the utterances within a window is the final score, as shown in equation \ref{score}, where $FCNN$ is the full-connected neural network. The algorithm details are shown in the Algorithm \ref{MIE_model}.

\begin{equation}
\begin{aligned}
    s[i] = max(FCNN(f[i])) \\
    y=sigmoid(s[i])
    \label{score}
\end{aligned}
\end{equation}


\begin{algorithm}
\label{MIE_model}
\caption{The MIE model}
\begin{algorithmic}
\Procedure{TheMIEProcedure}{}
\State \textbf{Input:} statement ${X}$
\State \textbf{Output:} final score $y$
\For{${X}_{i}$ in $X$}

// Encoder Module
\State $F[i]=BiLSTM[{X}_{i}]$
\State $d=\sum_{i=1}^n atttention weight \cdot F[i]$

// Matching Module
\State ${q}_{c}=\sum softmax({d}_{c},{F}_{c}) \cdot {F}_{c}$
\State ${q}_{s}=\sum softmax({d}_{s},{F}_{s}) \cdot {F}_{s}$

// Aggregation Module
\State $f[i]=concat[{q}_{c},{q}_{s}]$

// Score Module
\State $s[i]=max(FCNN(f[i])$
\State $y=sigmoid(s[i])$
\EndFor
\EndProcedure
\end{algorithmic}
\end{algorithm}

\subsection{KBQA module Based on User-constructed KB}
We accomplish a KBQA module that can answer users' natural language questions. It utilizes the user-constructed KG as a KB, which is stored in the graph database Neo4j. Neo4j supports the KBQA service with Cypher query statements as the search SQL to search answers in the database.
Next, we will elaborate on the technical architecture, the data collection method, and the KBQA implementation details.

\subsubsection{Technical Architecture}
\begin{figure}[htbp]
    \centering
    \includegraphics[width=\linewidth]{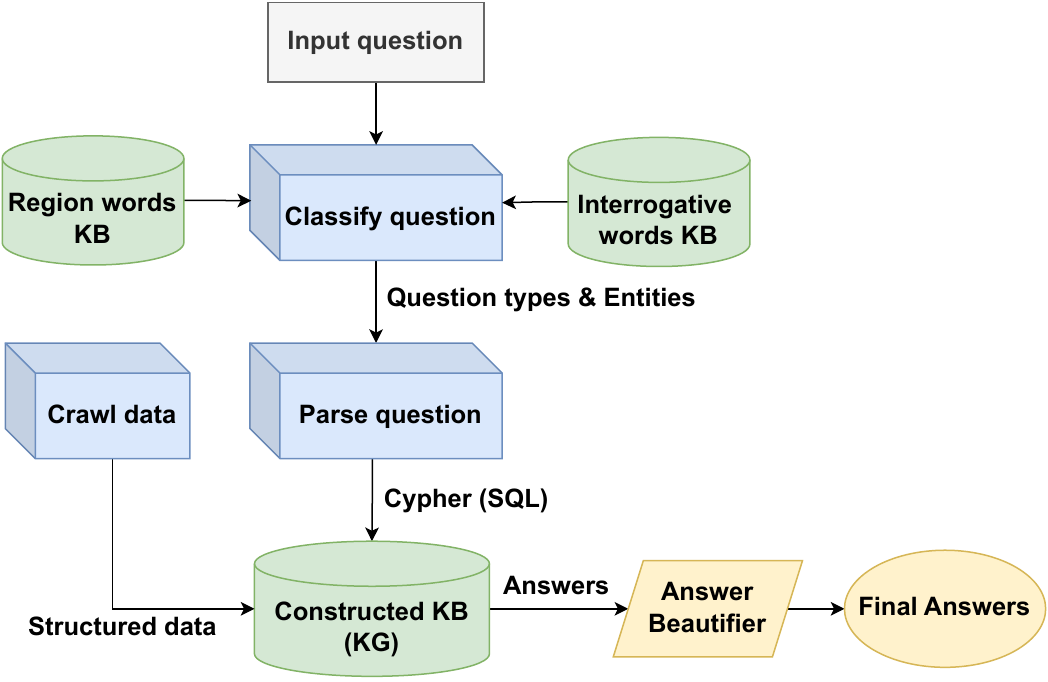}
    \caption{Technical Architecture}
    \label{Architecture}
\end{figure}

The technical architecture of the KBQA module is illustrated in Fig. \ref{Architecture}.  First, the question should be input by the user. Next, the question can be classified to obtain the question types and entities involved in it based on the region words and interrogative words KB. After that, according to the question types and entities, the question can be parsed to acquire the corresponding Cypher statements of the question. They can be used to query the Neo4j database, which stores the KG constructed by the crawled structured data, to obtain the answers. Finally, the answer beautifier can embellish the results returned by the KB according to pre-defined answer templates to obtain the final answers.

\subsubsection{Data Crawling}
\label{Crawling}
Initially, the Urllib library in Python is utilized to request
the HTML file of a certain website based on the URL of the main webpage. Then we parse the URLs of the HTML file to obtain more URLs of the classified information, which we can continue to crawl further information. 
Next, we parse the crawled files to acquire the basic information of the medical fields~\cite{zhan2023debiasing}, which are divided into numerous types of knowledge afterward.


\subsubsection{Classification of Question}
\begin{figure}[htbp]
    \centering
    \includegraphics[width=\linewidth]{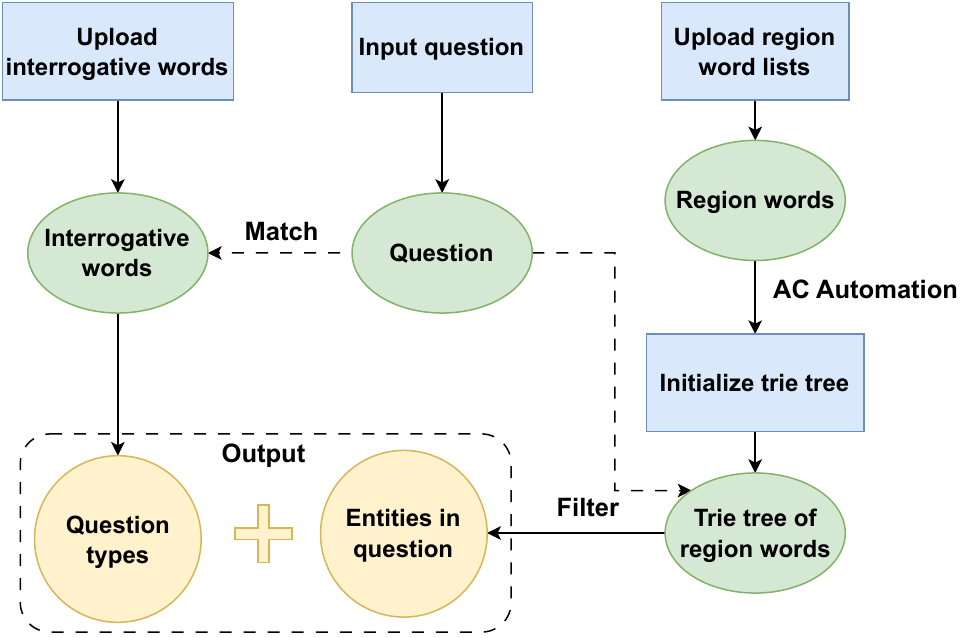}
    \caption{Question classification}
    \label{question_classification}
\end{figure}

In the classification stage, we need to obtain the question types and the entities by utilizing traditional rule-based matching algorithms and string-matching methods from the question input by a user. The whole process is illustrated in Fig. \ref{question_classification}. 
First, we need to utilize interrogative words in order to acquire the question types. The examples of altogether 13 types are shown in Table \ref{Interrogative}. 
The different kinds of interrogative words can be used to classify the input questions into different categories by matching the questions. If an interrogative word is in the question, then the question is the type of the type interrogative word belongs to. Notably, a question can belong to several question types. 
\begin{table}[htbp]
\caption{Interrogative word examples}
\label{Interrogative}
\begin{tabular}{@{}ll@{}}
\toprule
\textbf{Types} & \textbf{Word Examples}                         \\ \midrule
Symptom                            & Phenomenon,manifestation             \\
Cause                              & Cause, how, reason                             \\
Complications                      & Occurring together, accompanying \\
Food                               & Diet, drinking, supplements                    \\
Drug                               & Medicine, capsule, drug                        \\
Prevention                         & Prevent, avoid, escape                         \\
Lasting time                       & Period, how long, how many days                \\
Cure way                           & How to treat, how to heal, therapy             \\
Cure probability                   & Likelihood, curable, chances                   \\
Susceptible population             & Susceptible, infected, get                     \\
Check                              & Find out, check, measure out                   \\
Cure                               & Treat, cure, heal                              \\
Belong section                     & Belong to, what section, section               \\ \bottomrule
\end{tabular}
\end{table}
After that, region words of different entity types can be utilized to obtain the entities in question. The examples of region words are shown in Table \ref{region}. Due to the huge amount of region words, we utilize the Aho-Corasick (AC) algorithm to accelerate filtering, which is classical in multi-pattern string matching problems using AC automation. AC algorithm is a Knuth-Morris-Pratt (KMP) algorithm implemented on a trie tree to complete multi-pattern string matching, which aims to obtain all possible positions of all the pattern strings $P_1, P_2, P_3,\cdots, P_m$ in consecutive texts $T_1,\cdots, T_n$ for multiple pattern strings. 
The trie tree utilizes the common prefixes of strings to enhance efficiency by reducing the overhead of query time because once the tree is built, it can be queried many times. Finally, the question types and the entities
from the question we acquired in the classification stage are conducive to parsing the question in the next stage.

\begin{table}[htbp]
\caption{Region words types and examples}
\label{region}
\begin{tabular}{@{}ll@{}}
\toprule
\textbf{Types} & \textbf{Word Examples}                          \\ \midrule
Check                       & Body layer photography, Static imaging          \\
Department                  & Psychology, Gynecology, Otolaryngology          \\
Disease                     & Acromegaly, High arched foot                    \\
Drug                        & Brain and Blood Capsules, Ma Ren Pill           \\
Food                        & Sea shrimp and tofu, Red Pepper \\
Producer                    & Changke, Solnit                                 \\
Symptom                     & Low blood pressure, Holiday heart syndrome      \\ \bottomrule
\end{tabular}
\end{table}
\subsubsection{Parse and Search Question}
In this section, the KBQA module produces the appropriate Cypher query statements based on the classified question types. 
Each question type corresponds to one Cypher query template. Notably, each question may be converted to several Cypher statements as it may involve several entities.
Then Cypher statements are executed in the Neo4j database storing the KB constructed previously.
After that, the KB will return the raw results corresponding to the question.
Finally, the answer beautifier module of the KBQA will call the related reply template to embellish the raw answer according to the related question type, and then return the final answer of the question to the user.

\section{Results}
To better demonstrate the functions of our system, we utilize our method and model to construct the KG and the KBQA module based on the structured data crawled from the medical field. In this section, primarily, we present the KB scale constructed by the user. After that, we evaluate the performance of our model to construct KG by medical audio data. Finally, we illustrate the supported QA types of the KBQA module.

\subsection{Scale of KG Constructed by Structured Data}
We crawled from the medical website (jib.xywy.com) to collect structured data in the medical domain used to construct a medical KG.
The entity types of the KG consist of check items, department, disease, drug, food, producer, and symptom, with altogether about 33,000 entities.
The relation scale of KG is altogether about 230,000 relations.

\subsection{Evaluation of KG Constructed by Audio Data}
\subsubsection{The Results of VAD Model}
We applied the Librispeech \cite{7178964} dataset to evaluate the results in the VAD task. LibriSpeech is a corpus containing 1000 hours of English speech in 16 kHz. The audio is derived from people reading books from the LibriVox project and has been carefully preprocessed. Trained with the Librispeech dataset, the final accuracy of the validation set is 97.42\%, which indicates that our trained VAD model can distinguish non-speech and speech sections effectively.

\subsubsection{The Evaluation of SD Model}
\begin{figure}[htbp]
\centering{\includegraphics[width=7cm]{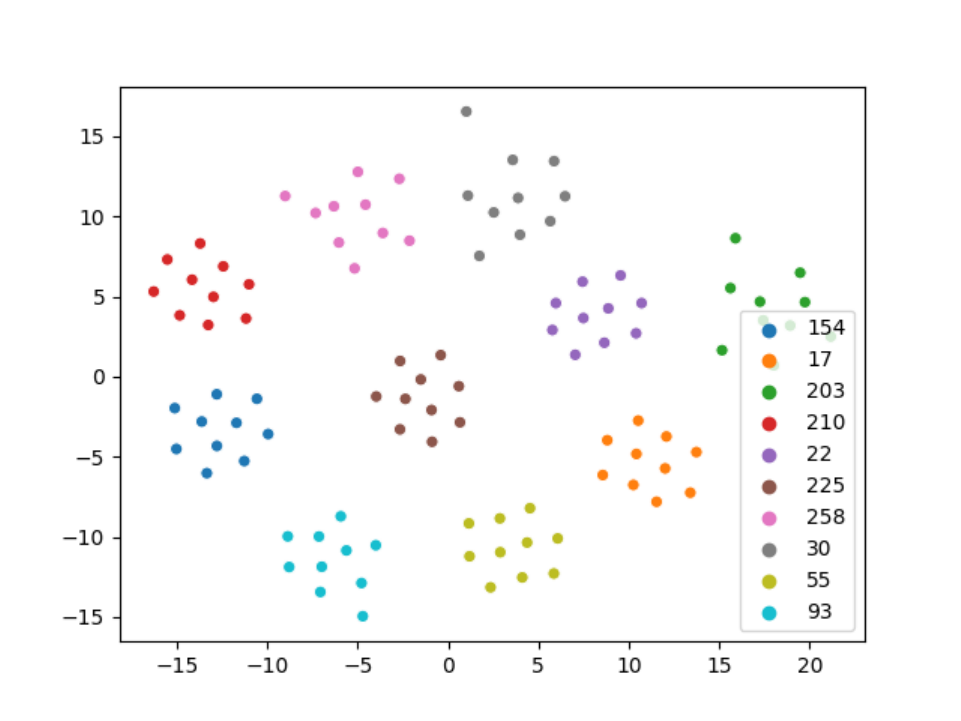}}
\caption{The classification result of ten speakers based on the GE2E model}
\label{ge2eresult}
\end{figure}

To prove the effectiveness of the SD model, we used LibriSpeech and VoxCeleb \cite{Nagrani_2017} datasets to train and validate the GE2E model. The VoxCeleb is a large-scale dataset for speaker identification. This data is collected from over 1,200 speakers of different accents, ages, and ethnicities. After training, the final EER (Equal Error Rate) is 10.58$\%$, proving that our model can classify different speakers in the audio. To better illustrate the classification results of the SD model,  we randomly pick ten speakers from LibriSpeech and visualize the results, as shown in Fig. \ref{ge2eresult}. The audio clips of each speaker are well-classified, indicating the effectiveness of our model. 

\subsubsection{The Results of MIE Model}
The doctor-patient dialogue dataset generated by Zhang et al. \cite{zhang-etal-2020-mie} is used to train and test the MIE model. This dataset uses the dialogues between patients and doctors from medical websites, and the labels are manually annotated. 

To verify the effectiveness of our method, we compare the MIE method with baselines from \cite{zhang-etal-2020-mie}.

\noindent
\textbf{Plain-Classifier} The classifier extracts features based on Bi-LSTM and self-attention mechanism to generate vectors. Then, the vectors are used to train the classifier. 

\noindent
\textbf{MIE-Classifier} This classifier utilized the MIE model architecture. The MIE-Classifier treats cutt c and cutt s directly as qc and qs, which is different from the situation in the MIE method.

\begin{table}[htbp]
\begin{center}
\caption{The Results of the MIE Model}
\begin{tabular}{cccc}
\hline
                 & P              & R              & F1             \\ \hline
Plain-Classifier & 61.34          & 52.65          & 56.08          \\
MIE-Classifier   & 71.87          & 56.67          & 61.78          \\
MIE              & \textbf{78.46} & \textbf{72.85} & \textbf{74.18} \\ \hline
\end{tabular}
\label{mieresult}
\end{center}
\end{table}

The prediction results of the full label "Category:Item-Status" are presented in Table \ref{mieresult}. The MIE method overperforms all the baselines, proving that the trained MIE model can precisely extract critical information from the medical dialogues.

\subsection{Supported QA Types of KBQA}
The KBQA module constructed by crawled data supports altogether 18 types of questions, which are shown in Table \ref{QA_types}.
The QA examples are shown in Table \ref{examples}.

\begin{table}[htbp]
\caption{Supported KBQA Types Examples}
\centering
\label{QA_types}
\begin{tabular}{@{}p{.4\linewidth}p{.6\linewidth}@{}}
\toprule
\textbf{Question Types}                 & \textbf{Question example}                                \\ \midrule
Disease symptoms                        & What are the symptoms of breast cancer?                  \\
Possible diseases according to symptoms & What should I do if I have a runny nose lately?          \\
Disease causes                          & Why do I suffer from insomnia?                           \\
Complications of disease                & What are the complications of insomnia?                  \\
Foods that diseases need to avoid       & What should people who have insomnia not eat?            \\
Food recommended for diseases           & What to eat if you have insomnia?                        \\
Disease need to avoid certain food      & Who is better off not eating honey?                      \\
Benefits of food for disease            & What are the benefits of goose meat?                     \\
Medicine to take for disease            & What medications should I take for liver disease?        \\
Disease prevention                      & What can I do to prevent insomnia?                       \\            \\
Disease Vulnerable Groups               & Who is susceptible to hypertension?                      \\
Disease description                     & Disease description Diabetes                             \\ \bottomrule
\end{tabular}
\end{table}

\begin{table}[htbp]
\caption{KBQA Examples}
\label{examples}
\begin{tabular}{@{}p{.35\linewidth}p{.65\linewidth}@{}}
\toprule
\textbf{Question}                                      & \textbf{Answer}                                                                                                                                                                                                                                                              \\ \midrule
What should I do to treat hypertension?       & Hypertension can try the following treatments: medication; surgery; supportive therapy                                                                                                                                                                              \\
Who is susceptible to hypertension?           & People who are susceptible to hypertension include: people with a family history of hypertension, poor lifestyle habits, and lack of exercise                                                                                                                       \\
What should people who have insomnia not eat? & Foods to avoid for insomnia include: doughnuts; mussels; lard                                                                                                                                                                                                       \\\bottomrule
\end{tabular}
\end{table}

\begin{figure*}[ht]
\centering
\includegraphics[width=\linewidth]{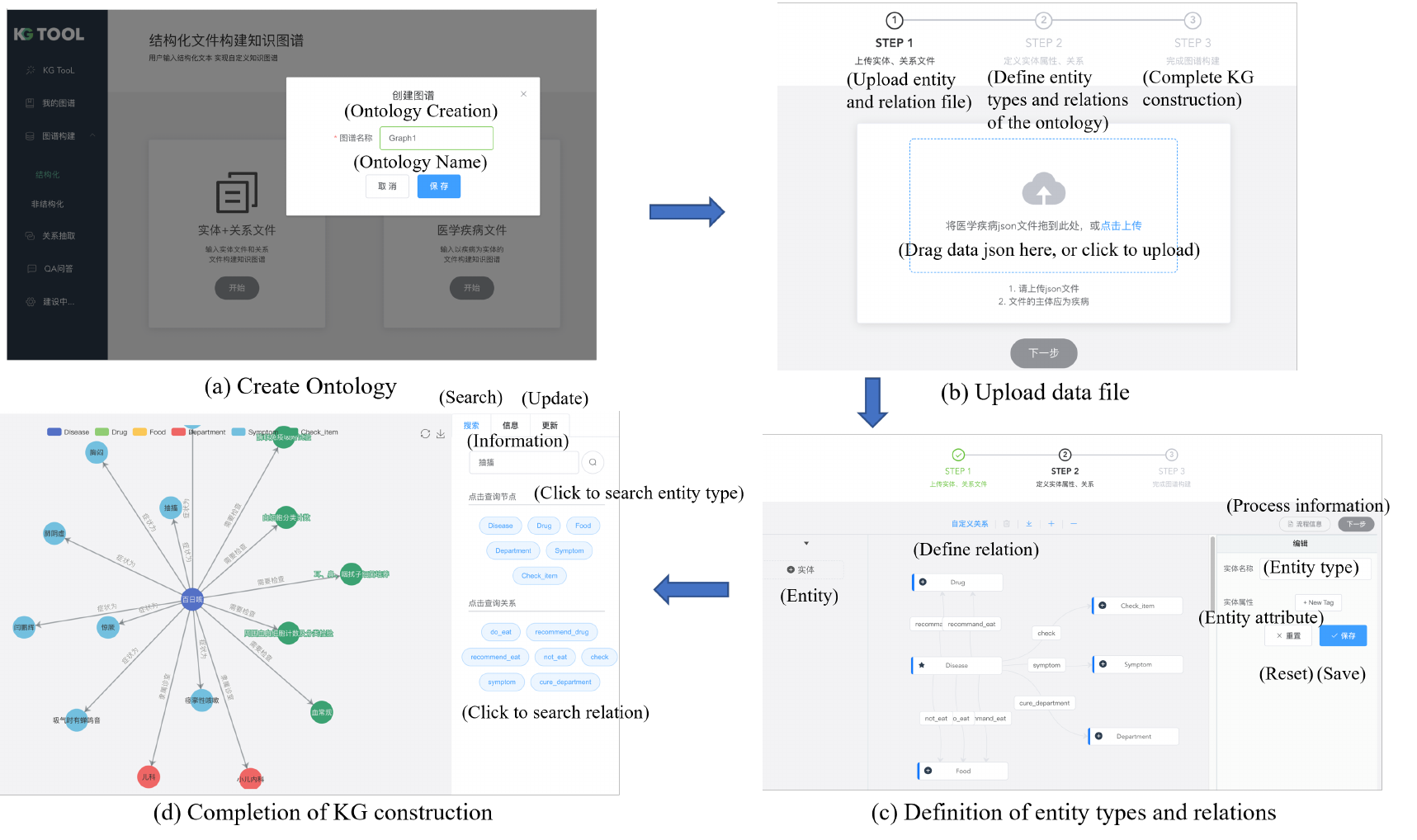}
\caption{KG construction procedures}
\label{KG construction}
\end{figure*}

\section{System Presentation}
In this section, first, we will introduce the technical route of the system. Next, we will illustrate the KG construction procedures and the ways to retrieve information in the system. After that, we will introduce the management of different KG versions. Finally, we will illustrate the KBQA module.
\subsection{Technical Route}
This whole system is developed using a front-end and back-end isolation strategy, with data being transferred between them using the HTTP protocol and the Restful API. The front-end frameworks used are Vue and ECharts, and the back-end framework used is Django. Additionally, we utilize graph database Neo4j to store KG data, and relational database SQLite to store the metadata of each KG.

\subsection{KG Construction Based on Structured Data}

\subsubsection{Construction Procedures}
The KG construction flow is illustrated in Fig. \ref{KG construction}.
Primarily, you need to create an ontology by defining its name, which is used to distinguish different versions of KGs. If the name does not exist in the database used to store the meta information of ontologies, it will be stored in the database. Otherwise, the system will prompt the duplicate name and you should change a different name. 
Next, you can upload the structured data file of the KG in JSON format. 
After that, you need to define the entity types and relations of the ontology by dragging entity type boxes and relation lines on the webpage. Additionally, you can delete unwanted entity types and relations. What's more, you can add a nickname and entity properties to the ontology.
Finally, the system will complete KG construction in the next step. Then the system will display some of the entities and relations randomly of the KG. 
As illustrated in subfigure (d) in Fig. \ref{KG construction}, different colors represent different entity types, and the arrows represent the relations between the entities. The right panel displays the entities and relations defined in the ontology.
\begin{figure*}[ht]
\centering
\includegraphics[width=\linewidth]{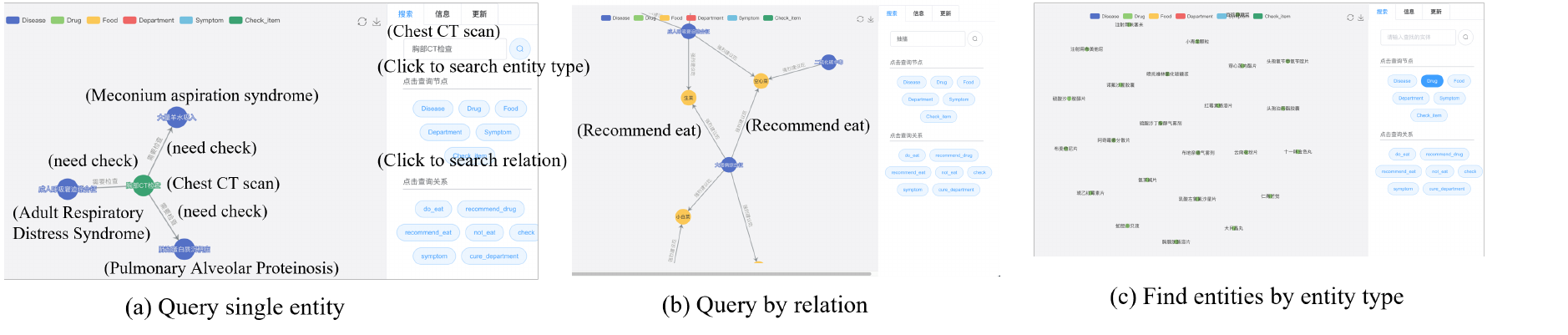}
\caption{Different query types}
\label{query}
\end{figure*}
\subsubsection{Information Retrieval}

We can retrieve information on nodes and relations after constructing the KG. First, we can query a single entity by searching its name in the right panel, as depicted in subfigure (a) in Fig. \ref{query}. What's more, we can retrieve all entities in an entity type or all graphs containing a relation by clicking the entity type and relation directly in the right panel, as shown in subfigures (b) and (c) in Fig. \ref{query}.
Moreover, we can retrieve the attributes of each node and relation by clicking it directly, which will be shown in the right panel directly.
Finally, undesirable nodes and relations can be deleted forever by clicking the trash button on the right panel. 

\subsection{KG Construction Based on Unstructured Data}
We can construct a KG from unstructured data by either uploading a recorded MP3 conversation audio or recording it in real time.
\subsubsection{Upload the recorded audio}
\begin{figure}[htbp]
    \centering
    \includegraphics[width=\linewidth]{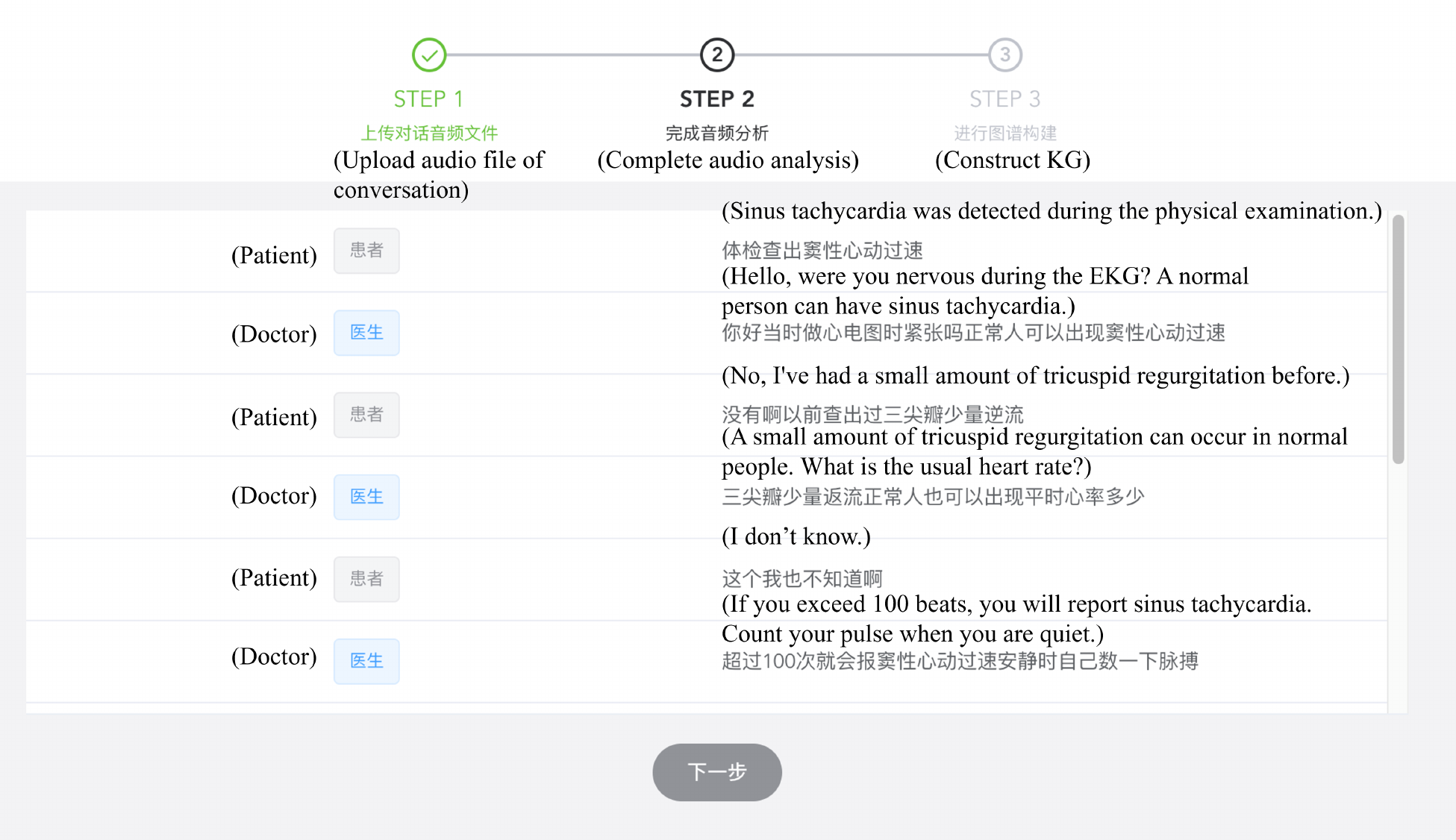}
    \caption{The converted results from speech to text}
    \label{ASR}
\end{figure}

Initially, enter the user's basic information and submit the conversation audio, of which the format is MP3 and the sample rate is 16K. The system will analyze the audio after uploading and convert the voice into text, as illustrated in Fig. \ref{ASR}.

\begin{figure}[htbp]
    \centering
    \includegraphics[width=\linewidth]{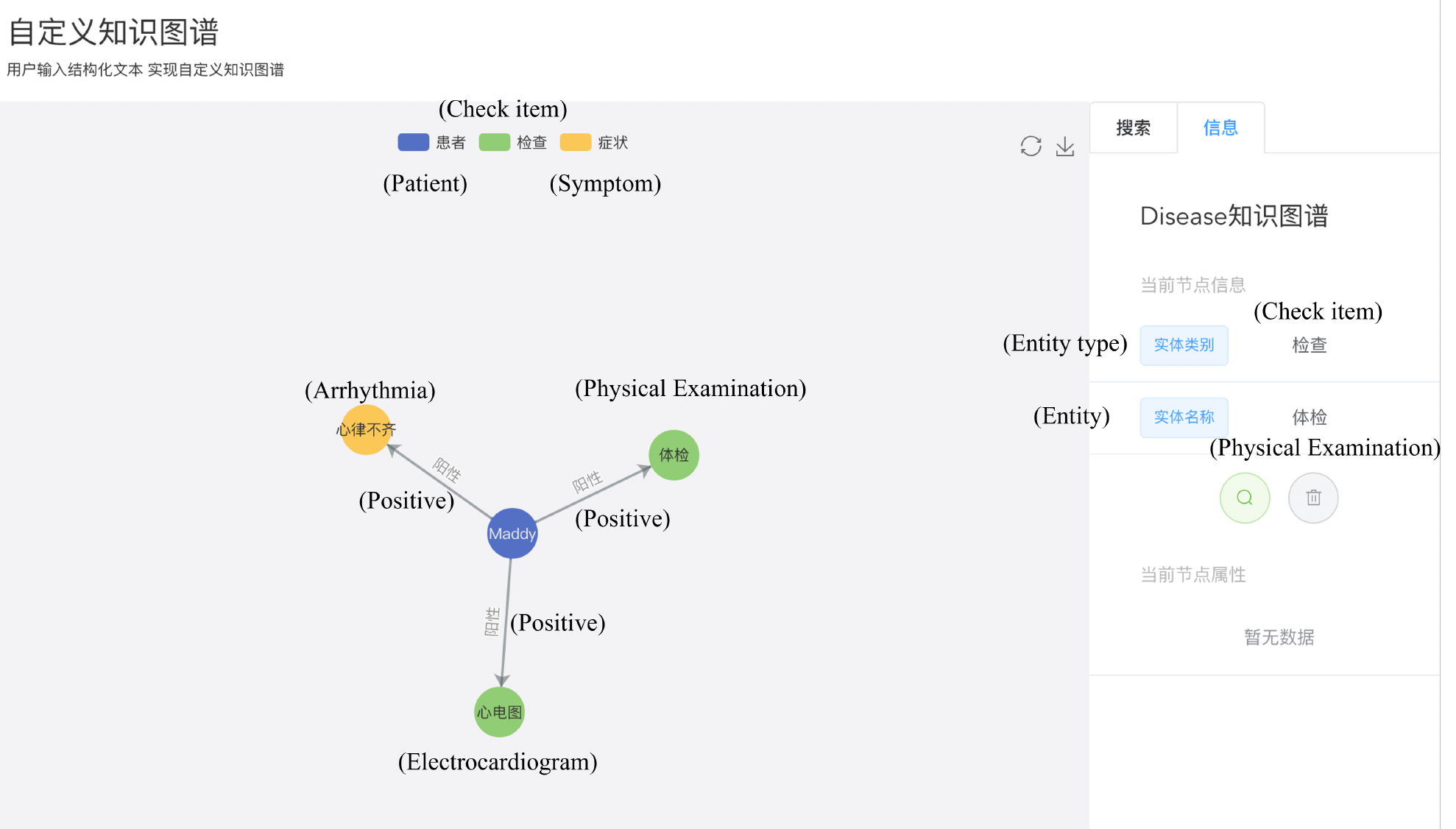}
    \caption{Return analyzed results based on KG}
    \label{uploadresult}
\end{figure}

After clicking the next step button, the information will be extracted from the text by the MIE model.
Next, the system will return the analyzed outcomes and present the results based on the KG once the analysis is accomplished, as depicted in Fig. \ref{uploadresult}.

\subsubsection{Record dialogues in real time}
\begin{figure}[htbp]
    \centering
    \includegraphics[width=\linewidth]{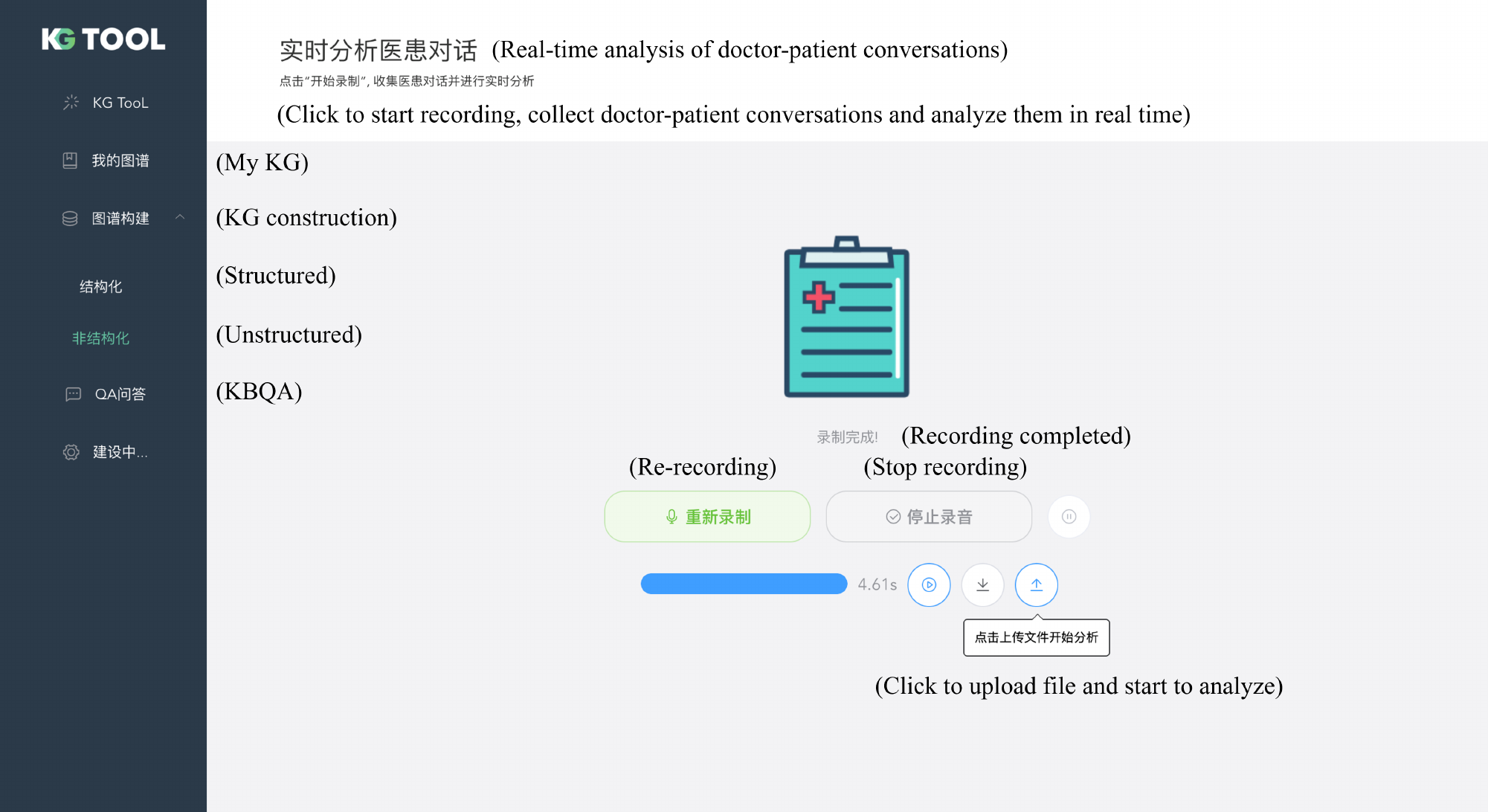}
    \caption{Record dialogue audio in real time}
    \label{record}
\end{figure}

Initially, enter the basic information of the users and begin to record the conversation in real time. At any point throughout the recording process, customers have the option to stop and restart the recording. They may also download the audio in the WAV file or play it again. The recording webpage is shown in Fig. \ref{record}.

\begin{figure}[htbp]
    \centering
    \includegraphics[width=\linewidth]{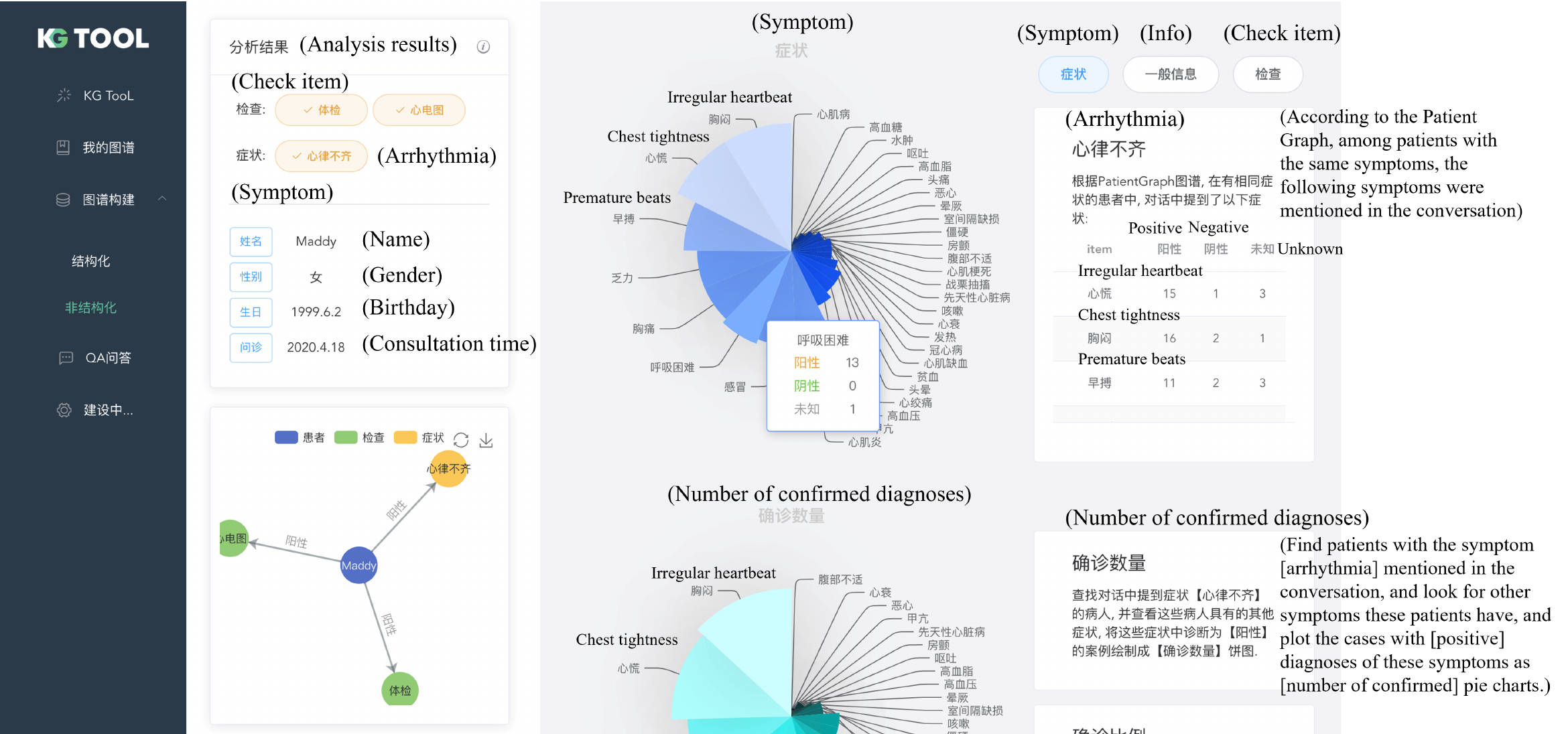}
    \caption{The extraction results}
    \label{recordresult}
\end{figure}

After uploading the audio, the system will analyze it and use the MIE model to extract crucial information from the conversation. The system then displays the results as demonstrated in Fig. \ref{recordresult}.

Next, the system mainly presents two types of information. First, on the left side, it will demonstrate the information of patients. Moreover, the extraction results of the dialogue will also be visually illustrated in the KG format. Secondly, the analysis of the data extracted from the patient based on KG is shown on the right side. 
It displays some crucial details on other patients who have the same symptoms as the present user.
For instance, in Fig. \ref{recordresult}, it is determined that  the user ``Maddy'' exhibits symptoms of arrhythmia. As a result, the right section lists statistical data about other individuals who have arrhythmia. It may aid in a more thorough analysis of the patient by the doctor.

\subsection{KG Version Management}
\begin{figure}[htbp]
\centering
\includegraphics[width=\linewidth]{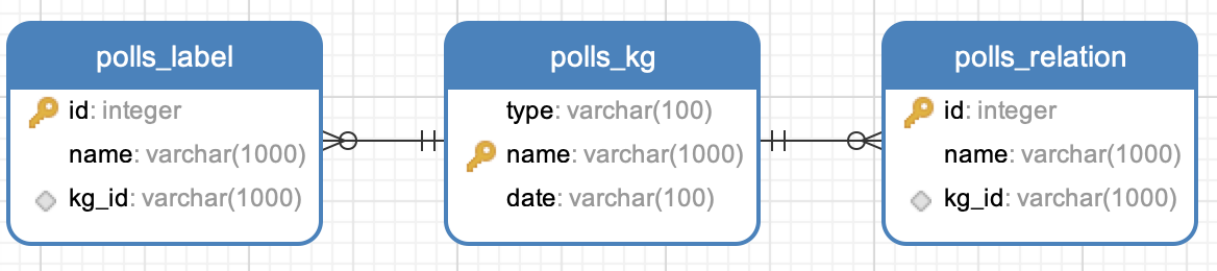}
\caption{Database schema}
\label{数据库模型}
\end{figure}

We can store the constructed KGs in a graph database, i.e. Neo4j, which can be viewed whenever needed. Additionally, we can add new data in the same format to the constructed KGs whenever we want. To achieve the function, we store the metadata of each KG in a relational database, i.e., SQLite, of which the database schema is shown in Fig. \ref{数据库模型}. The database model consists of three tables: KG, Lable, and Relation.
The primary key of the KG table is ``name'', which represents the name of a KG. That's why the name of each KG cannot be the same. It also serves as the foreign key of the Label and Relation tables, which are used to store the entity types and relations in an ontology respectively. The KG table is in a one-to-many relationship with the Label and Relation tables.

\begin{figure}[htbp]
\centering
\includegraphics[width=\linewidth]{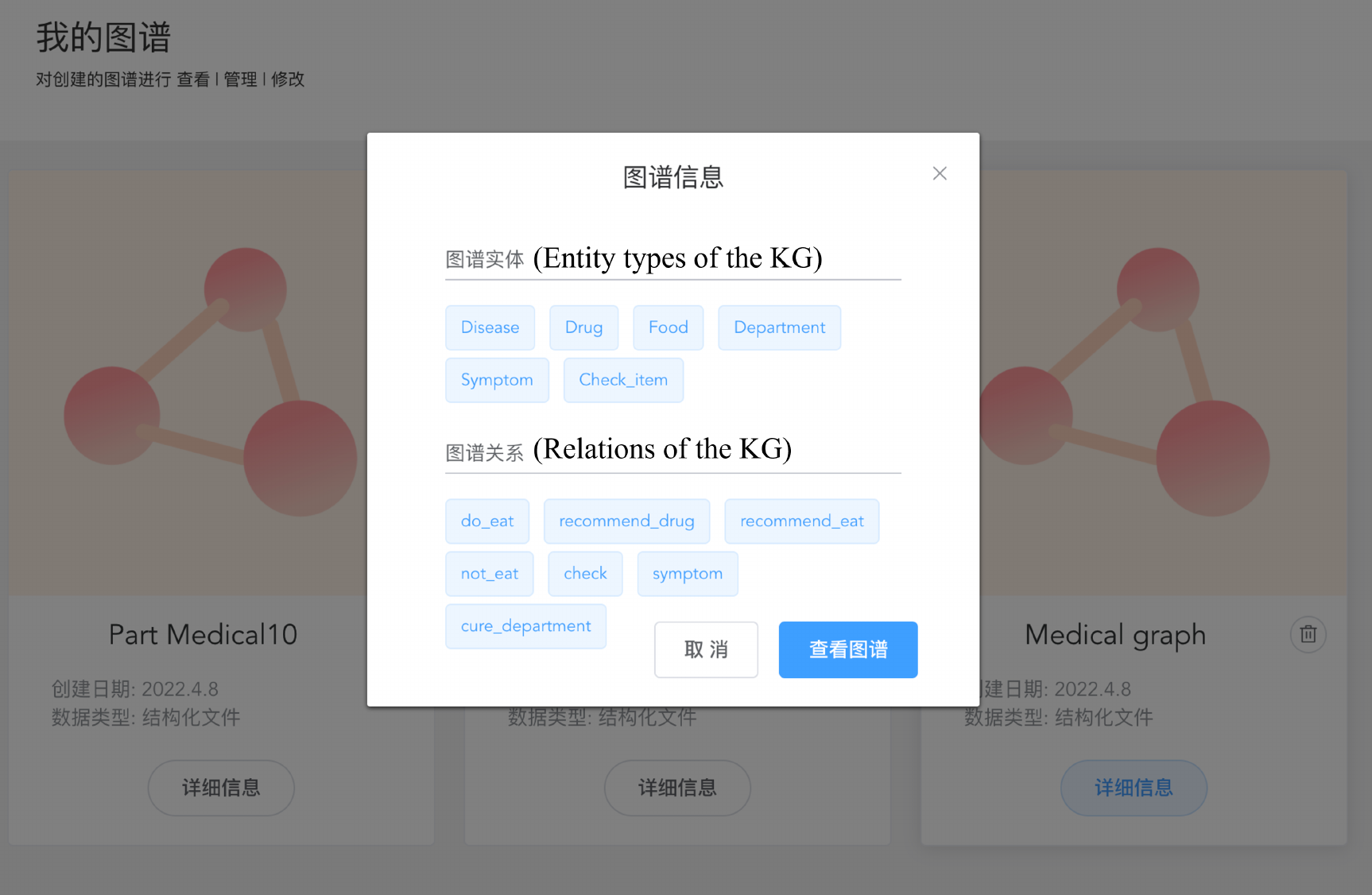}
\caption{KG version control and information}
\label{图谱信息}
\end{figure}

As illustrated in Fig. \ref{图谱信息}, the system can present all the KGs and their specific information, such as their creation time and data type, according to the metadata of the KGs stored in the database. Moreover, we can choose a KG to view its entity types and relations. You can manage the KGs by deleting undesirable KGs, of which all entities and relations will be deleted in the Neo4j database and all metadata will be deleted in the SQLite.

\subsection{KBQA Module}

\begin{figure}[htbp]
    \centering
    \includegraphics[width=\linewidth]{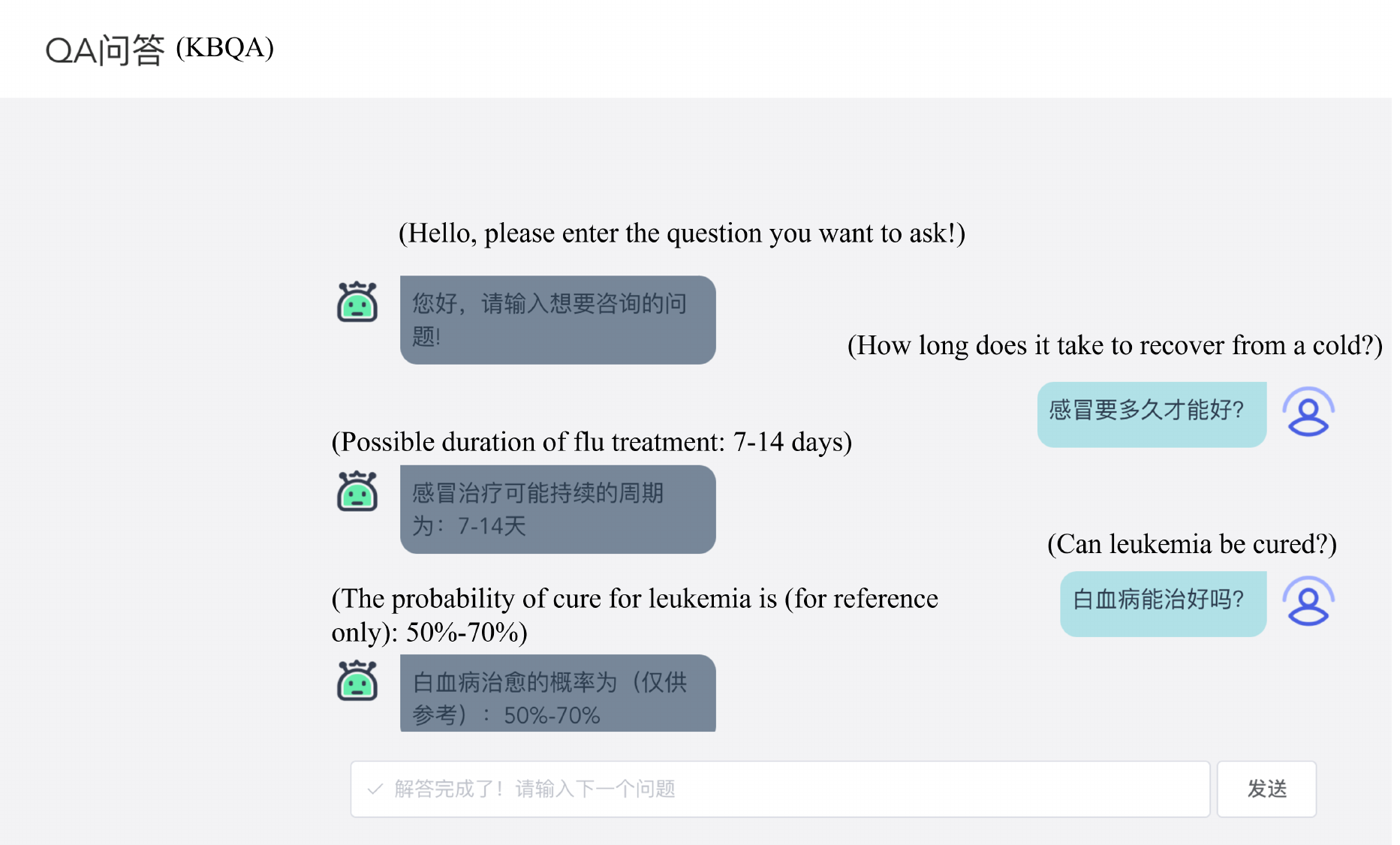}
    \caption{KBQA examples}
    \label{QA}
\end{figure}

As depicted in Fig. \ref{QA}, this KBQA system can receive the questions of users and then return accurate and concise answers to users. It is conducive to promoting access to health knowledge for ordinary people by applying the KBQA system to the medical area. Moreover, it can assist doctors to diagnose efficiently, which considerably alleviates the medical pressure on society.

\section{Conclusion}
In this article, we propose an intelligent KG construction and application platform SAKA and prove the automatic KG construction method feasible on the SAKA.
It offers a user-friendly and intuitive KG construction technique, which only requires data upload and button operation to achieve semi-automatic KG construction, application, and management, in contrast to other platforms requiring expert knowledge and computing ability. 
Moreover, we propose the AGIE method to construct KGs by extracting semantic information from structured and audio data. We also evaluate the effectiveness of AGIE on serval datasets.
Moreover, we also develop a KBQA module based on the KGs constructed by users.

Nevertheless, several potential limitations of the SAKA platform still exist. Primarily, the scalability of SAKA might be a problem when encountering large-scale KGs. This is an aspect we aim to address in our future work to ensure optimal performance even under heavy data loads. The handling of noisy data was also identified as a critical challenge. We are also planning on implementing more sophisticated error-handling mechanisms to tackle this issue. Finally, the need for handling domain-specific knowledge more efficiently was brought to our attention. Although our current platform permits users to customize entity types and relations, it is necessary to accommodate complex domain-specific scenarios more effectively. Future developments will aim to extend the platform's capabilities by integrating domain-specific models and rules.

\backmatter
\section*{Declarations}
\bmhead{Sources of Funding}
This work was supported by the National Key RD Program of China under Grant (No. 2020YFB1707803).
\bmhead{Conflict of interest}
The authors have no relevant financial or non-financial interests to disclose.
\bmhead{Informed consent}
Not applicable
\bmhead{Ethical approval}
Not applicable
\bmhead{Acknowledgments}
Part of this paper is extended from a conference paper originally presented at the IEEE ICEBE 2022 conference. The authors also would like to thank the conference organizers for their invitation to extend the paper.  

\bibliographystyle{sn-basic}
\bibliography{ref}

\end{document}